\documentclass[letter, 10 pt, conference]{ieeeconf}
\usepackage[noadjust]{cite}
\usepackage{amsmath}
\usepackage{url}  
\usepackage{graphicx}
\usepackage{epsfig}
\usepackage{epstopdf}
\usepackage{subfigure} 
\IEEEoverridecommandlockouts
\usepackage{multirow}
\usepackage{makecell}
\DeclareMathOperator*{\argminA}{arg\,min}
\title{\LARGE \bf
Road Crack Detection Using Deep Convolutional\\ Neural Network and Adaptive Thresholding
}
\author{Rui Fan$^{1*}$, Mohammud Junaid Bocus$^{2*}$, Yilong Zhu$^{3}$,\\ $\ \ \ \ \ \ $ Jianhao Jiao$^{1}$, Li Wang$^{4}$, Fulong Ma$^{3}$, Shanshan Cheng$^{4}$, Ming Liu$^{1}$
\thanks{$^1$R. Fan, J. Jiao and M. Liu are with the Robotics and Multi-Perception Laboratory in Robotics Institute at the Hong Kong University of Science and Technology, Hong Kong.}
\thanks{$^2$M. J. Bocus is with the Visual Information Institute at the University of Bristol, Bristol, United Kingdom.}
\thanks{$^3$Y. Zhu and F. Ma are with Unity-Drive Technology Inc, Shenzhen, China.}
\thanks{$^4$L. Wang and S. Cheng are with the National Engineering Research Center of Road Maintenance Technologies, Beijing, China.}
\thanks{$^*$These two authors contributed equally to this work and therefore are joint first authors. Corresponding author: Rui Fan. Email: rui.fan@ieee.org.}
}
\begin{document}

\maketitle
\thispagestyle{empty}
\pagestyle{empty}

\begin{abstract}
Crack is one of the most common road distresses which may pose road safety hazards. Generally, crack detection is  performed by either certified inspectors or structural engineers. This task is, however, time-consuming, subjective and labor-intensive. In this paper, we propose a novel road crack detection algorithm based on deep learning and adaptive image segmentation. Firstly, a deep convolutional neural network is trained to determine whether an image contains cracks or not. The images containing cracks are then smoothed using bilateral filtering, which greatly minimizes the number of noisy pixels. Finally, we utilize an adaptive thresholding method to extract the cracks from road surface. The experimental results illustrate that our network can classify images with an accuracy of $\boldsymbol{99.92\%}$, and the cracks can be successfully extracted from the images using our proposed thresholding algorithm. 
\end{abstract}
\section{INTRODUCTION}
\label{sec.introduction}
Road damages in the form of cracks may reduce the road performance and pose potential road safety hazards \cite{Fan2018c}. Every year, government bodies across the globe allocate funds to enhance the quality of their road networks \cite{oliveira2009automatic}. Road safety should be taken very seriously and authorities are fully aware of the need for suitable road inspection and maintenance techniques \cite{Fan2018}. Crack detection is an essential part of road maintenance systems, and it has attracted growing interest from researchers in this field over the past few years \cite{Nguyen2009}. 
Traditional manual road crack detection approaches are known to be very time-consuming, dangerous, labor-intensive and subjective \cite{Fan2018b}. Therefore, the slow and subjective traditional methods have been gradually replaced by automated crack detection systems which provide fast and reliable analysis in intelligent transportation systems (ITS) \cite{Fan2019}. Automated crack detection systems can effectively assess the quality of the road surfaces and help  governments plan and prioritize the maintenance of the road network, thereby keeping the roads in good condition and extending their service life \cite{Oliveira2013}. 

With the development of image analysis techniques, road crack detection and recognition have been widely investigated over the past few decades \cite{Oh1998, Petrou1996, Huang2006, Gavilan2011}. The traditional framework for crack detection consists of defining a variety of gradient features using gradient filters, such as Sobel \cite{Ozgunalp2017, Fan2016},  for each image pixel, and then using a binary classifier to determine whether an image pixel is part of a crack region or not \cite{Oh1998}. In early methods, such as \cite{Kaseko1993} and \cite{Li2008}, the authors used threshold-based approaches to find crack regions based on the assumption that a pixel lying in a crack area is consistently darker than others \cite{Oliveira2013, Fan2018g}. Furthermore, many researchers \cite{Huang2006, Gavilan2011, Tanaka1998, Oliveira2008} tried to suppress the inference of noise by considering additional local features, such as the mean and the standard deviation of an image region. However, these methods are still very sensitive to noise because only the brightness features are taken into consideration.

In recent years, some novel algorithms, such as minimal path selection (MPS) \cite{Jiao2019, Avila2014, Amhaz2015}, minimum spanning tree (MST) \cite{Zou2012, Fernandes2014} and crack fundamental element (CFE) \cite{Tsai2012, Tsai2014}, have been proposed to improve the existing crack detection approaches. In addition, Hu and Zhao \cite{Hu2010} proposed a crack detection algorithm based on local binary patterns (LBP), whereas the authors of \cite{Salman2013} utilized Gabor filter for the same purpose. In \cite{Zou2012}, an automated crack detection algorithm based on a tree structure, referred to as CrackTree, was introduced. Moreover, Oliveira et al. \cite{Oliveira2013, Oliveira2014} utilized a comprehensive set of image analysis algorithms to detect and characterize cracks from road pavements. Although the above-mentioned algorithms have been widely used in crack detection and they perform well on high-quality datasets \cite{Zou2012, Oliveira2014, Varadharajan2014}, it is important to note that these algorithms are not accurate enough to distinguish cracks from the complex background in low-quality images. 

\begin{figure*}[t]
	\begin{center}
		\centering
		\includegraphics[width=0.93\textwidth]{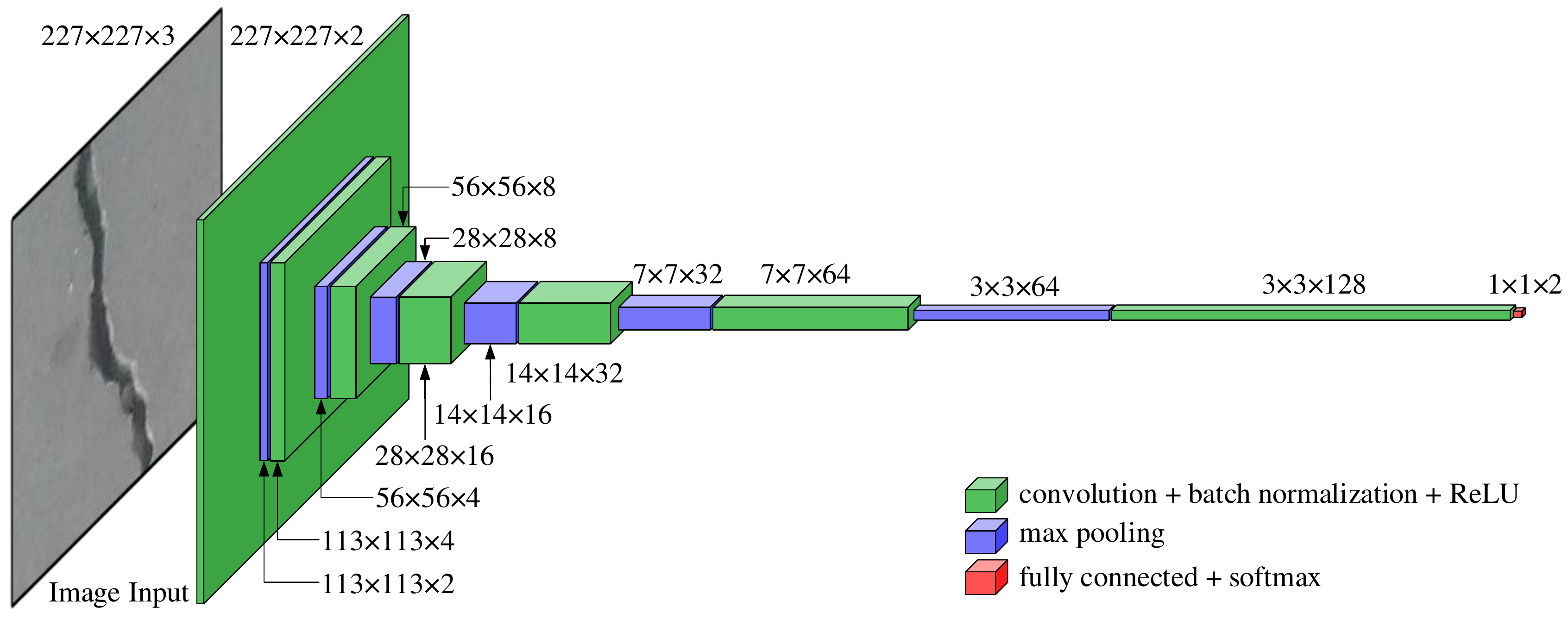}
		\centering
		\caption{The structure of the proposed deep neural network for image classification. }
		\label{fig.cnn}
	\end{center}
\end{figure*}
\begin{figure}[t]
	\begin{center}
		\centering
		\includegraphics[width=0.42\textwidth]{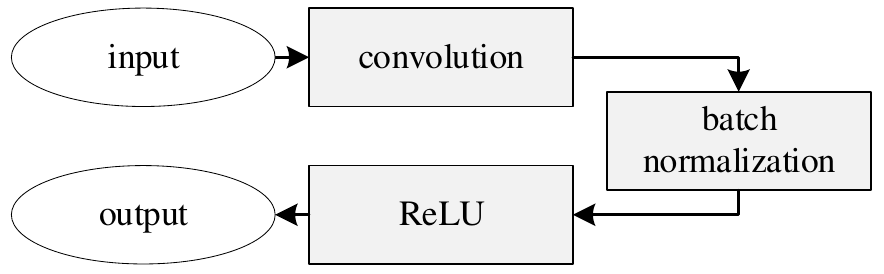}
		\centering
		\caption{The structure of the green block in Fig. \ref{fig.cnn}. }
		\label{fig.block_unit}
	\end{center}
\end{figure}

Furthermore, some machine learning-based crack detection approaches \cite{Roth2016, Ciresan2012, Ciresan2013, Krizhevsky2012, Zhang2015, Kivinen2014} have been proposed in recent years, and the features produced by neural network are very likely to replace the local features utilized in traditional methods \cite{Zhang2016}. For example, restricted Boltzmann machine (RBM) anto encoder and their variants are capable of detecting cracks, when the training samples are limited \cite{Xu2018}. In addition,  deep convolutional neural networks (DCNNs) are popular for feature-learning and supervised classification \cite{LeCun2015}. Zhang et al. \cite{Zhang2016} trained a neural network to determine whether the patches in an images contain cracks or not.  Hence in this paper, we build on the recent successful application of deep neural network to image classification and train a convolutional neural network (CNN) to find the images that contain cracks. Then, we present a novel thresholding method to extract cracks from classified color images.

The remainder of this paper is structured as follows: Section \ref{sec.methodology} introduces the proposed crack detection algorithm. In Section \ref{sec.experimental_results}, we present our experimental results and discuss the performance of the proposed method. Finally, Section \ref{sec.conlcusion} concludes the paper and provides some recommendations for future work. 
\section{Methodology}
\label{sec.methodology}
The proposed crack detection method consists of two steps: image classification and image segmentation. For notational convenience, images showing the presence and absence of cracks are referred to as positive and negative images, respectively. Firstly, an image is classified as either positive or negative using a deep convolutional neural network. The positive images are then processed using an adaptive thresholding method. The cracks in the positive images can therefore be extracted. The rest of this section gives a detailed description of these two steps.

\subsection{Image Classification}
\label{sec.image_classification}
The structure of the proposed deep convolutional neural network is shown in Fig. \ref{fig.cnn}, where ReLU represents a rectified linear unit, which is the most popular activation function for deep neural networks, due to its better performance than both sigmoid function  and hyperbolic tangent function  in terms of training and evaluation \cite{LeCun2015}. A CNN is generally considered as a hierarchical feature extractor \cite{Zhang2016}. A convolutional layer performs a convolution operation on the image input and passes the extracted features to the next layer \cite{LeCun2015}.  
Batch normalization is then performed on the output of the convolutional layer, whereby the extracted features are normalized by adjusting and scaling the activations. \cite{Ioffe2015}. The structure of the green block in Fig. \ref{fig.cnn} is shown in Fig. \ref{fig.block_unit}. Max pooling downsamples the input representations \cite{LeCun2015}, whereas the softmax function translates a vector into a probability distribution. Finally, a fully connected layer computes the score of each class and infers the category of the input image \cite{Zhang2016}. Therefore, the proposed network is also referred to as a fully connected network (FCN). More details on the training process are provided in Section \ref{sec.experimental_results}.
\subsection{Image Segmentation}
\label{sec.image_segmentation}
\begin{figure}[t]
	\begin{center}
		\centering
		\subfigure[]{
			\includegraphics[width=0.13\textwidth]{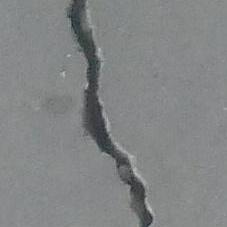}
			\label{fig.img}
		}
		\subfigure[]{
			\includegraphics[width=0.13\textwidth]{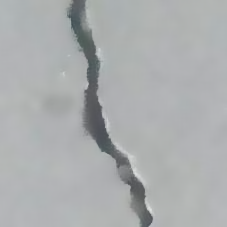}
			\label{fig.fb}
		}
			\subfigure[]{
		\includegraphics[width=0.13\textwidth]{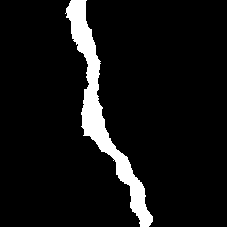}
		\label{fig.res}
	}
		\centering
		\caption{Bilateral filtering and image segmentation; (a) original positive image;  (b) filtered positive image;  (c) segmentation result. }
		\label{fig.bilateral_filtering}
	\end{center}
\end{figure}
Since the images have already been classified using our proposed deep neural network, only the positive images are considered for processing in this subsection.  
Before performing image segmentation, we first utilize a bilateral filter \cite{Fan2018a, Fan2018e} to smooth the input images. Bilateral filter outperforms other image filters in terms of edge preservation \cite{Fan2018a}. A general expression for bilateral filtering is as follows: 
\begin{equation}
i_\text{bf}(u,v)=\frac{\sum_{x=u-\rho}^{u+\rho}\sum_{y=v-\rho}^{v+\rho}\omega_s(x,y)\omega_c(x,y)i(x,y)}{\sum_{x=u-\rho}^{u+\rho}\sum_{y=v-\rho}^{v+\rho}\omega_s(x,y)\omega_c(x,y)},
\label{eq.bf}
\end{equation}
where
\begin{equation}
\omega_s(x,y)=\exp\Bigg\{\frac{(x-u)^2+(y-v)^2}{{\sigma_s}^2}  \Bigg\},
\end{equation}
\begin{equation}
\omega_c(x,y)=\exp\Bigg\{\ \frac{(i(x,y)-i(u,v))^2}{{\sigma_c}^2} \ \Bigg\}.
\end{equation}
$i(x,y)$ represents the intensity of a pixel at $(x,y)$ in the input image. $i_\text{bf}(u,v)$ denotes the intensity of a pixel at $(u,v)$ in the filtered image. $\omega_s$ and $\omega_c$ are based on spatial distance and color similarity, respectively. Their values are controlled by two parameters $\sigma_s$ and $\sigma_c$, respectively. In our experiments, the values of $\sigma_s$ and $\sigma_c$ are set to  300 and 0.1, respectively. $\rho$ is set to 5. The filtered image is shown in Fig. \ref{fig.bilateral_filtering}.

\begin{figure}[!t]
	\begin{center}
		\centering
		\includegraphics[width=0.34\textwidth]{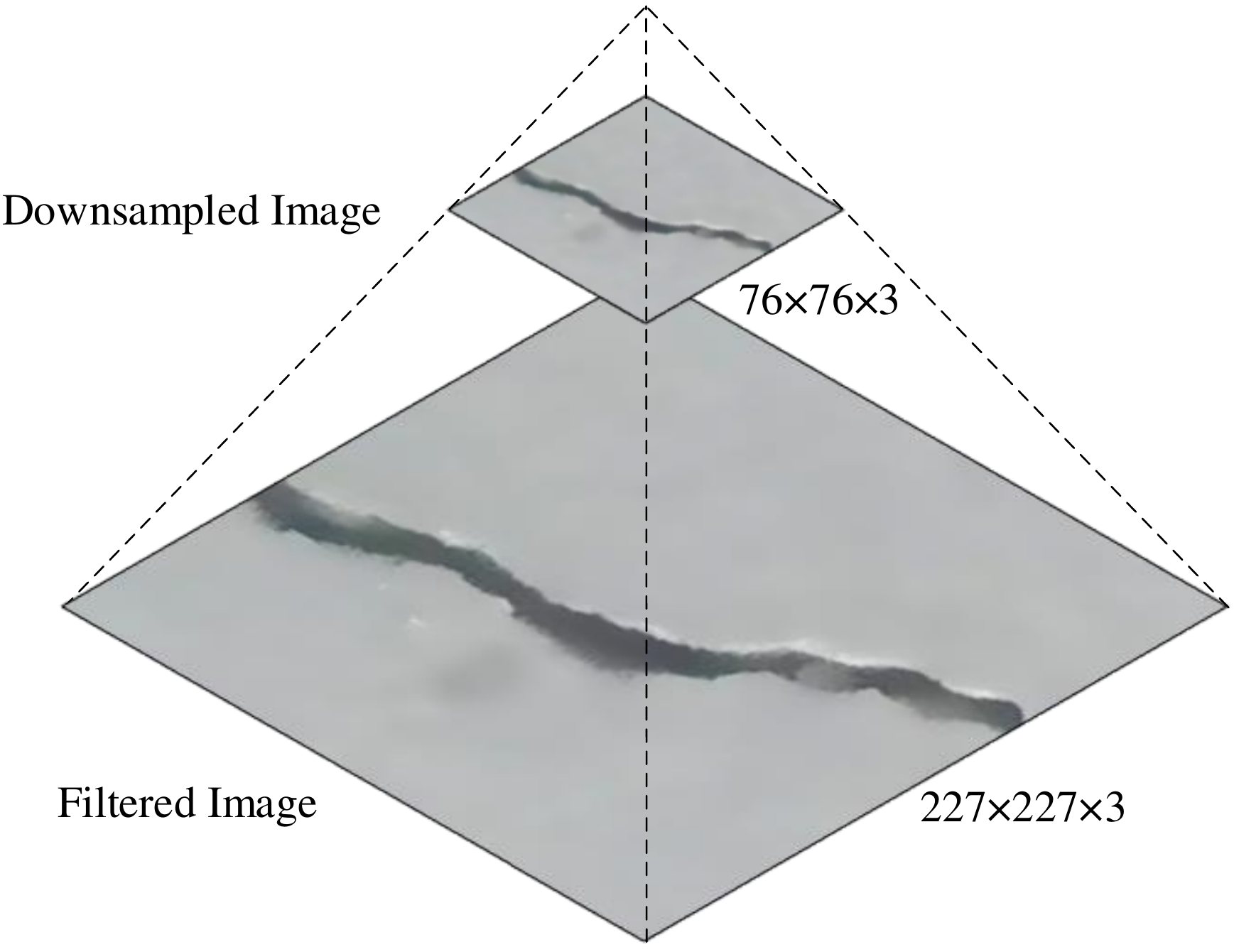}
		\centering
		\caption{Image downsampling. }
		\label{fig.downsampling}
	\end{center}
\end{figure}

To further reduce noise in the filtered image, the latter is downsampled  as shown in Fig. \ref{fig.downsampling}.  The downsampled image is approximately nine times smaller than the original filtered image and it is utilized as the threshold for image segmentation.  It is to be noted that the intensity of a pixel in the downsampled image is normalized. 

\begin{figure}[!t]
	\begin{center}
		\centering
		\includegraphics[width=0.43\textwidth]{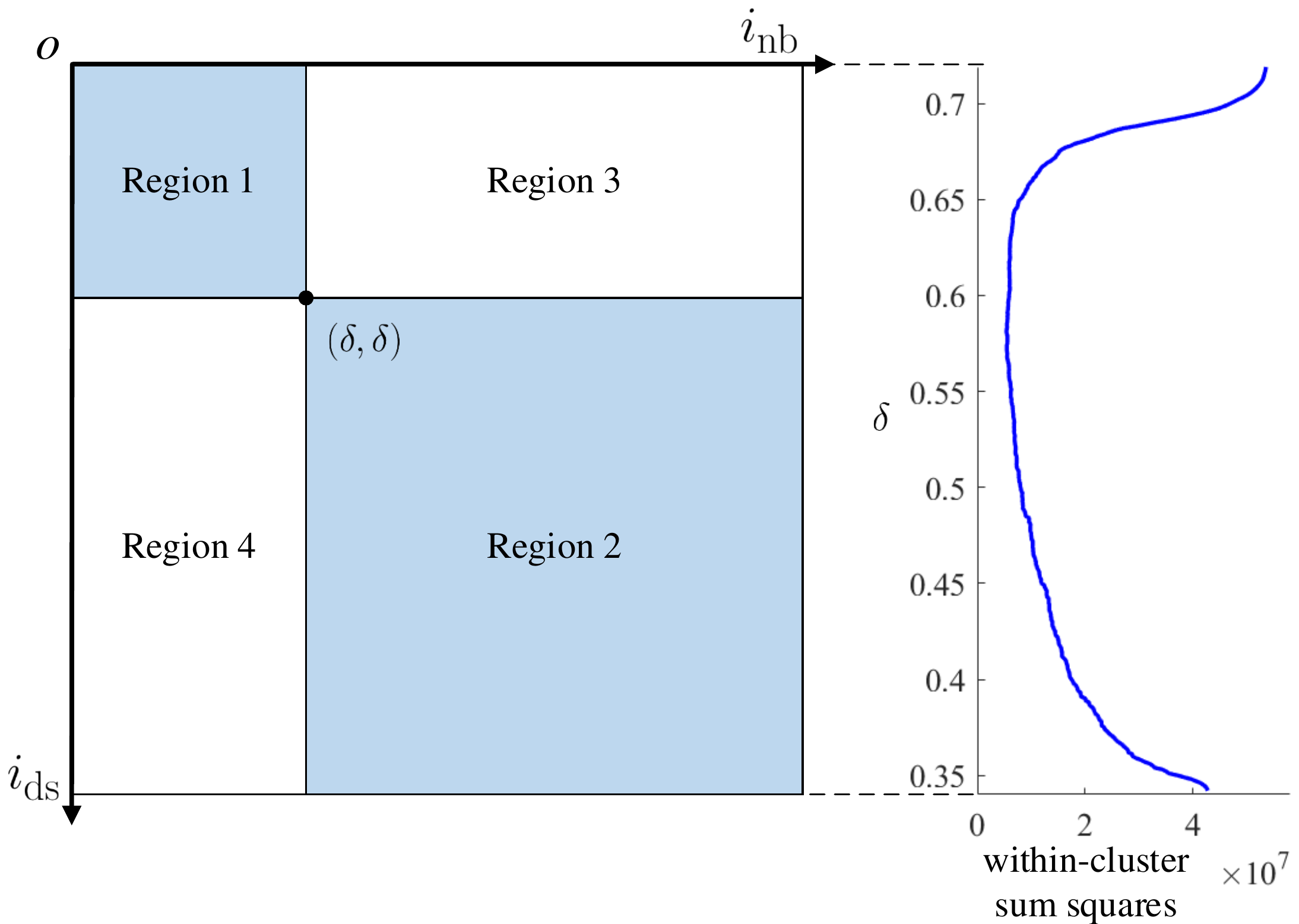}
		\centering
		\caption{2D histogram and the within-cluster sum squares with respect to different $\delta$. }
		\label{fig.thresholding}
	\end{center}
\end{figure}

The proposed thresholding method hypothesizes that the downsampled image is composed of two parts: foreground (cracks) and background (road surface), and they can be separated using one threshold $\delta$. Furthermore, we assume that a pixel lying in a crack area is consistently darker than others. 
To find the best threshold $\delta$, we formulate the thresholding problem as a 2D vector quantization problem, where each pixel $\boldsymbol{p}=[x,y]^\top$ and its neighborhood system $\boldsymbol{Q}=[\boldsymbol{q_p}_0, \boldsymbol{q_p}_1, \dots,  \boldsymbol{q_p}_n ]^\top$ provide a vector $\boldsymbol{m}=[i_\text{ds}, i_\text{nb}]^\top$, where $i_\text{ds}$  represents the intensity of $\boldsymbol{p}=[x,y]^\top$, and $i_\text{nb}$ denotes the mean intensity  of $\boldsymbol{Q}$. The vectors are stored in a 2D histogram, as shown in Fig. \ref{fig.thresholding}.  The relationship between $i_\text{ds}$ and $i_\text{nb}$ is as follows:
\begin{equation}
i_\text{nb}(u,v)=\sum_{x=u-\tau}^{u+\tau}\sum_{y=v-\tau}^{v+\tau}i_\text{ds}(x,y)-i_\text{nb}(u,v),
\label{eq.ds_nb}
\end{equation}
where $\tau$ dictates the size of the neighborhood system.  The threshold can be determined by partitioning the vectors into two clusters $\boldsymbol{S}=\{\boldsymbol{S}_1,\boldsymbol{S}_2\}$, where $\boldsymbol{S}_1$ and $\boldsymbol{S}_2$ correspond to the foreground and background, respectively. 

According to the Markov Random Field \cite{Boykov2001}, the intensity of a pixel which is not located near the boundary between foreground and background is similar to those of its neighbors in all directions.  Hence,  we search for the threshold along the principal diagonal of the 2D histogram using $k$-mean clustering \cite{Ahmad2007}. Given a threshold $\delta$, the 2D histogram can be divided into four regions, as shown in Fig. \ref{fig.thresholding}. Regions 1 and 2 store the vectors of foreground and background, respectively. On the other hand, regions 3 and 4 store the noisy vectors. In our method, the vectors in regions 3 and 4 are not considered into the clustering process. The best threshold is computed  by minimizing the within-cluster sum squares as follows:
\begin{equation}
\begin{split}
\argminA_{\boldsymbol{S}}\sum_{i=1}^{2}\sum_{\boldsymbol{m}\in\boldsymbol{S}_i}||\boldsymbol{m}-\boldsymbol{\mu}_i||^2,\\
\end{split}
\label{eq.k_mean}
\end{equation}
where $\boldsymbol{\mu}_i$ denotes the mean of the vectors in $\boldsymbol{S}_i$.  %Eq. \ref{eq.k_mean} can be rearranged as follows:
%\begin{equation}
%\begin{split}
%\argminA_{\boldsymbol{S}} \sum_{i=1}^2\sum_{\boldsymbol{m}\in\boldsymbol{S}_i}||\boldsymbol{m}||^2-n_{1}||\boldsymbol{\mu}_1||^2-n_{2}||\boldsymbol{\mu}_2||^2,
%\end{split}
%\label{eq.k_mean1}
%\end{equation}
%where $\boldsymbol{\mu}_1$ and $\boldsymbol{\mu}_2$ represent the means of vectors in regions 1 and 2, respectively. $n_{1}$ and $n_{2}$ denote the numbers of vectors in regions 1 and 2, respectively. 
The within-cluster sum squares with respect to different ${\delta}$ are shown in Fig. \ref{fig.thresholding}, and the corresponding segmentation result is shown in Fig. \ref{fig.res}, where the crack and road surface are in white and black, respectively. The performance of the crack detection algorithm is evaluated  in Section \ref{sec.experimental_results}.

\section{Experimental Results}
\label{sec.experimental_results}
In our experiments, the proposed deep neural network is trained on an NVIDIA GTX 1080 Ti GPU\footnote{https://www.nvidia.com/en-us/geforce/products/10series/geforce-gtx-1080-ti/}, which has 3584 CUDA cores and 11 GB GDDR5X memory. The GPU memory bandwidth is 484 GB/s. The training is implemented on Matlab R2018b. Our trained neural network is publicly available at: \url{https://github.com/ruirangerfan/road_crack_detection_net.git}. 

The dataset\footnote{http://dx.doi.org/10.17632/5y9wdsg2zt.1} utilized for training the proposed network was created by the researchers from Middle East Technical University. The dataset contains 40000 RGB images (resolution: 227$\times$227).  The number of positive and negative images are both 20000. 

\begin{figure*}[!t]
	\begin{center}
		\centering
		\includegraphics[width=1\textwidth]{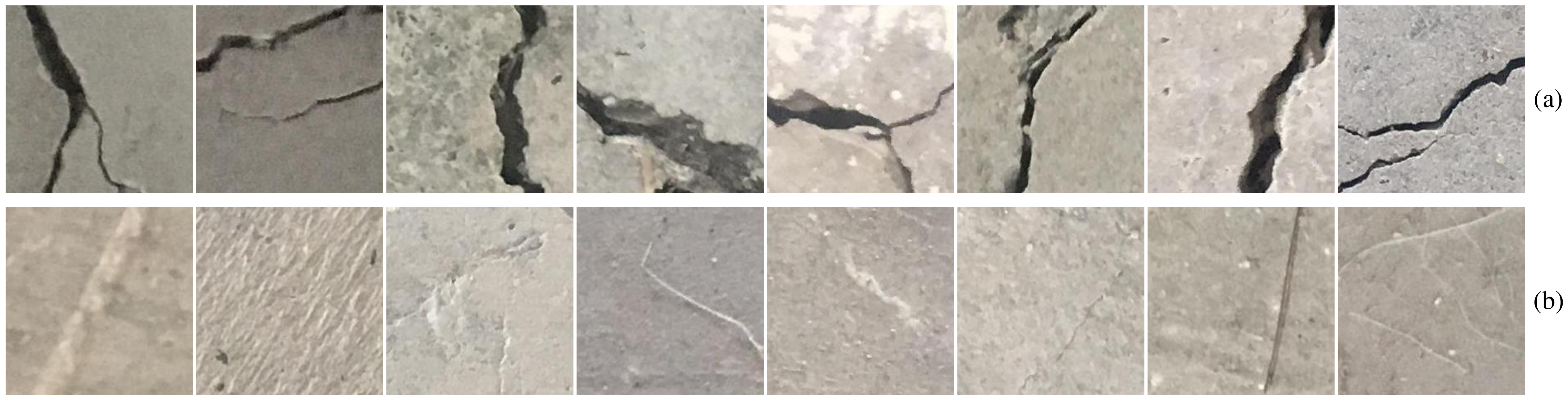}
		\centering
		\caption{Experimental results of image classification. (a) True positive images. (b) True negative images.}
		\label{fig.true}
	\end{center}
\end{figure*}

\begin{figure}[!t]
	\begin{center}
		\centering
		\includegraphics[width=0.48\textwidth]{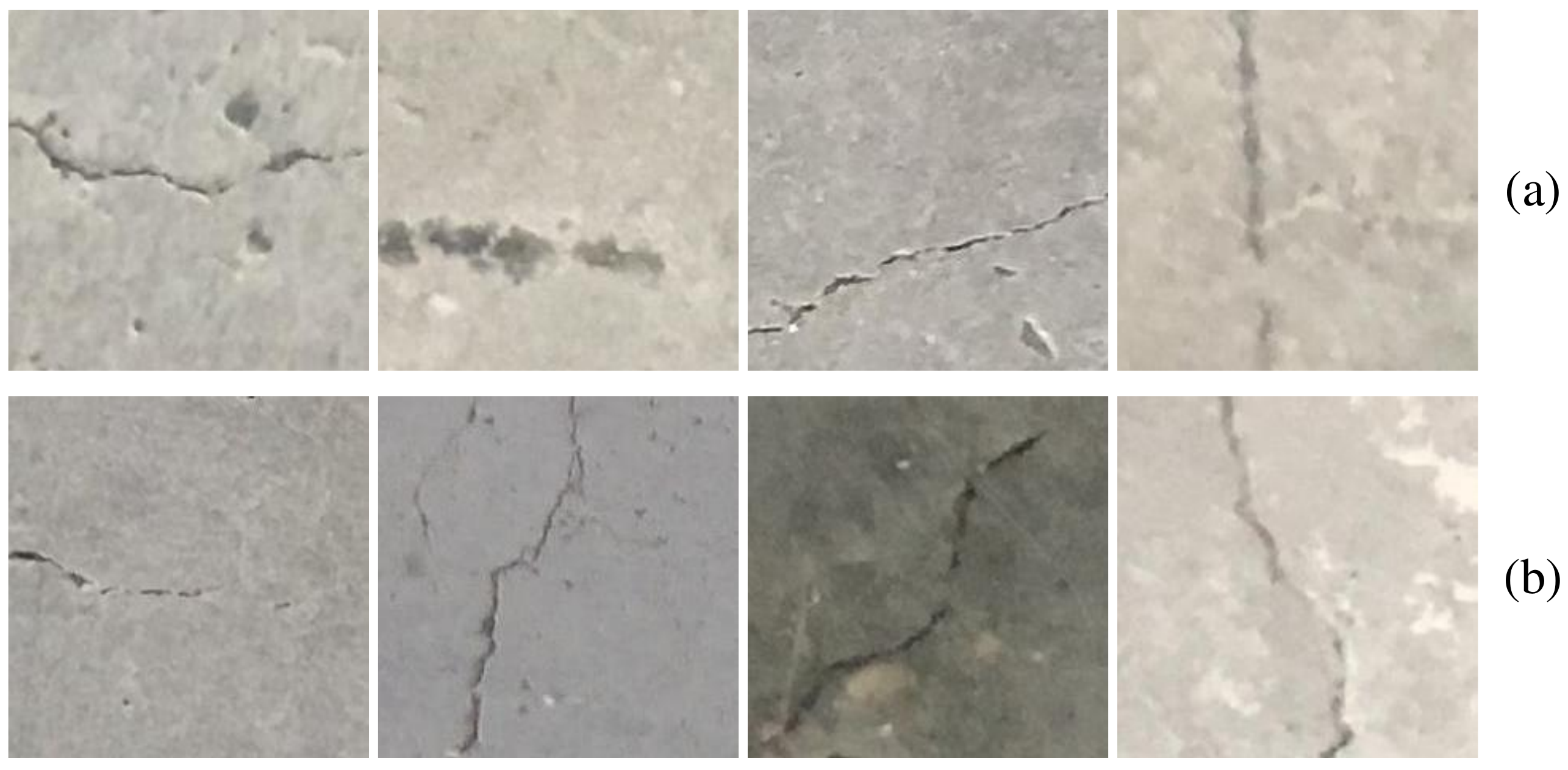}
		\centering
		\caption{The failed classification results. (a) False positive images. (b) False negative images.}
		\label{fig.fp_fn}
	\end{center}
\end{figure}

\begin{figure*}[!t]
	\begin{center}
		\centering
		\includegraphics[width=0.95\textwidth]{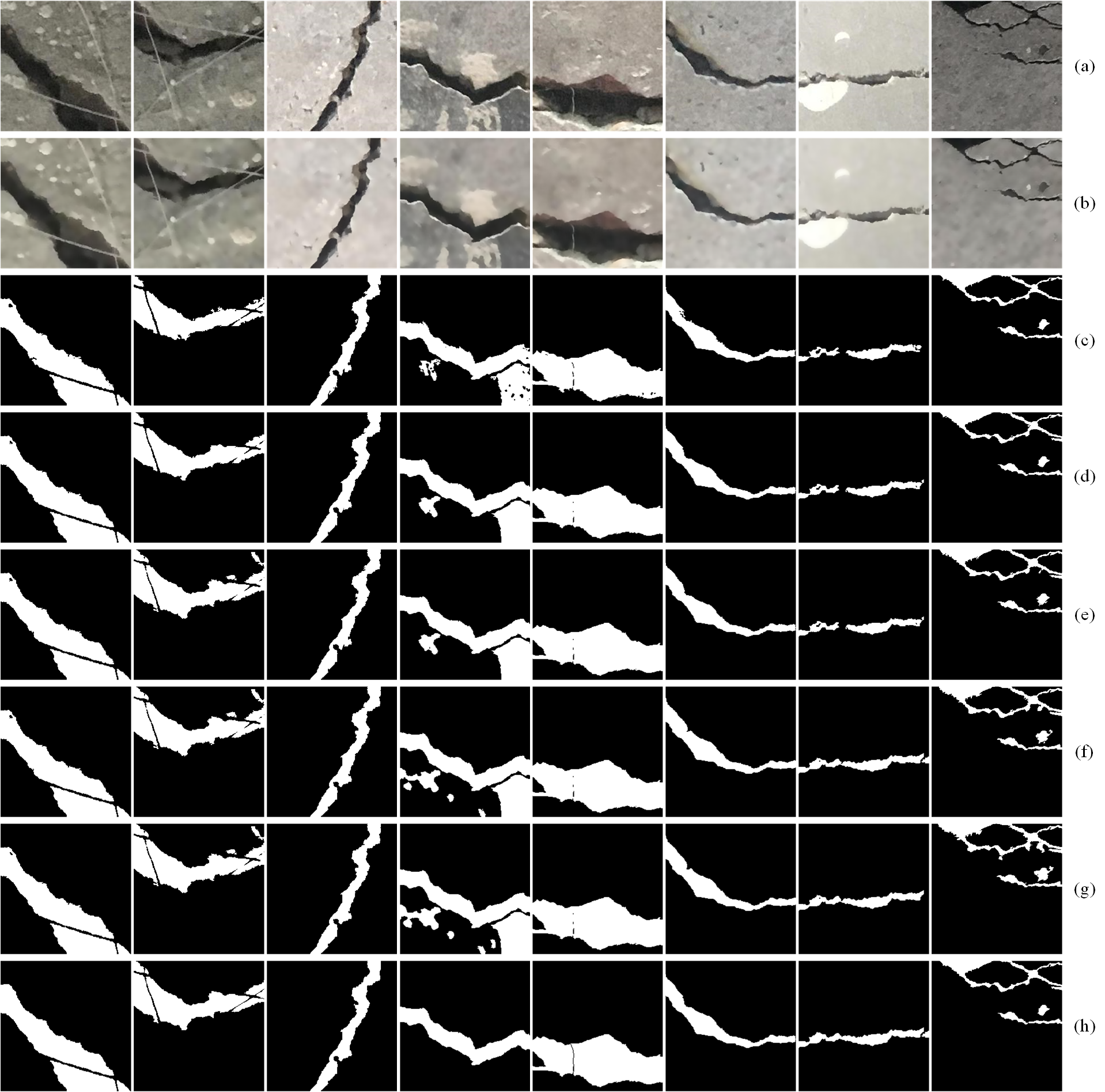}
		\centering
		\caption{Experimental results of image segmentation; (a) input images; (b) filtered images;  (c) results obtained using Otsu's thresholding method; (d)-(g) results obtained using the proposed method when $\tau$ is set to $1$, 2, 3 and 4, respectively; (h) ground truth.}
		\vspace{-1em}
		\label{fig.seg_res}
	\end{center}
\end{figure*}

In our practical experiments, we randomly select 15000 positive images and 15000 negative images from the dataset, to train the neural network. The rest of the images are utilized to evaluate the performance of the proposed approach. The initial learning rate, the maximum number of epochs and the validation frequency are set to 0.01, 16 and 60, respectively. The stochastic gradient descent with momentum (SGDM) is utilized as the optimizer, and the value of momentum is set to 0.9. 

To quantify the accuracy of our proposed image classier, we compute $n_\text{tp}$, $n_\text{fp}$, $n_\text{fn}$ and $n_\text{tn}$, which represent the number of testing images that are true positive, false positive, false negative and true negative, respectively. The precision, recall,  accuracy and F$_1$-measure can be computed using the following equation: 
\begin{equation}
	\text{precision}=\frac{n_\text{tp}}{n_\text{tp}+n_\text{fp}},
	\label{eq.pre}
\end{equation}
\begin{equation}
\text{recall}=\frac{n_\text{tp}}{n_\text{tp}+n_\text{fn}},
\label{eq.rec}
\end{equation}
\begin{equation}
\text{accuracy}=\frac{n_\text{tp}+n_\text{tn}}{n_\text{tp}+n_\text{tn}+n_\text{fp}+n_\text{fn}},
\label{eq.acu}
\end{equation}
\begin{equation}
\text{F}_1=2\frac{\text{precision}\cdot\text{recall}}{\text{precision}+\text{recall}}.
\label{eq.f1}
\end{equation}
The values of precision, recall, accuracy and F$_1$-measure achieved using the proposed network are $99.92\%$ (the number of false positive images and false negative images are both four). The true positive and true negative examples are shown in Fig. \ref{fig.true}, while the false positive and false negative results are shown in Fig. \ref{fig.fp_fn}. The image classification takes about 4.8 ms on an Intel Core i7-8700K CPU processed with a single core (3.7 GHz). 

Furthermore, we evaluate the performance of the image segmentation at pixel level.  Some experimental results of image filtering and segmentation are shown in Fig. \ref{fig.seg_res}. Since the dataset we use does not contain pixel-level ground truth, we manually label the crack areas in a set of images and use these ground truth images to quantify the performance of our proposed segmentation method. Moreover, we compare our method with Otsu's thresholding method \cite{Otsu1979} with respect to pixel-level precision, recall, accuracy and F$_1$-measure. The notations $n_\text{tp}$, $n_\text{fp}$, $n_\text{fn}$ and $n_\text{tn}$ in (\ref{eq.pre}), (\ref{eq.rec}) and (\ref{eq.acu}) represent the number of pixels that are true positive, false positive, false negative and true negative, respectively. The comparison between these two methods is shown in Table \ref{table.comparsion}. It is to be noted here that the crack areas with less than 100 pixels are ignored in our experiments. From Table \ref{table.comparsion}, it can be observed that our method outperforms Otsu's thresholding method in terms of precision, accuracy and F$_1$-measure, and the proposed segmentation method achieves the best performance when $\tau$ is set to 1. 

\begin{table}[!t]
	\begin{center}
		\vspace{0in}
		\footnotesize
		\caption{Comparison between our proposed method and Otsu's thresholding method. }
		\label{table.comparsion}
		\begin{tabular}{|c|c|c|c|c|}
			\hline
			Method & Precision & Recall & Accuracy & F$_1$-measure\\ 
			\hline
			Otsu's thresholding & 0.9590 & \textbf{0.9339} & 0.9848 & 0.9462\\ 
			Proposed ($\tau=1$) & 0.9774 & 0.9331 & \textbf{0.9870} & \textbf{0.9548}\\
			Proposed ($\tau=2$) & 0.9854 & 0.9246 & 0.9867 & 0.9541\\
			Proposed ($\tau=3$) & 0.9967 & 0.9046 & 0.9848 & 0.9484\\
			Proposed ($\tau=4$) & \textbf{0.9955} & 0.8952 & 0.9831 & 0.9427\\
			\hline
		\end{tabular}
	\end{center}
\end{table}

\section{Conclusion and Future work}
\label{sec.conlcusion}
A novel crack detection approach was proposed in this paper. The main novelties include a fully connected neural network for image classification and a $k$-mean clustering based image segmentation algorithm. Firstly, our neural network classified the input images as either positive (crack present) or negative (crack absent). The positive images were then processed using a bilateral filter, which not only minimized the number of noisy pixels but also preserved the edges between the cracks and road surface. Finally, the filtered images were downsampled, and an adaptive threshold was computed by minimizing the within-cluster sum squares. The cracks can therefore be detected by segmenting the filtered images using the adaptively determined threshold. The experimental results showed that the precision of the image classification is $99.92\%$ and the pixel-level segmentation accuracy is around $98.70\%$. 

Although the proposed image segmentation algorithm performs better than Otsu's thresholding method in terms of distinguishing between foreground (cracks) and background (road surface), some color images with a large number of noisy pixels cannot be properly segmented. Therefore, as a future work, a deep neural network can be trained to segment the positive images into a set of semantically meaningful regions, i.e., cracks and road surface. 

%Furthermore, we plan to utilize the proposed algorithm to detect other road damages, e.g., potholes, spalling, rutting, etc.

\section*{Acknowledgment}
\label{sec.ack}
This work is supported by grants from  the Research Grants Council of the Hong Kong SAR Government, China (No. 11210017 and No. 21202816) awarded to Prof. Ming Liu. 
This work is also supported by grants from the Shenzhen Science, Technology and Innovation Commission, JCYJ20170818153518789, and National Natural Science Foundation of China (No. 61603376) awarded to Dr. Lujia Wang.

\bibliographystyle{IEEEtran}

\end{document}